\documentclass[preprint,12pt]{elsarticle}

\usepackage{amssymb}
\usepackage{amsmath}
\usepackage{subcaption}

\usepackage{graphicx}
\usepackage{adjustbox}

\newcommand{\imgbox}[1]{%
  \adjustbox{width=0.22\textwidth,height=0.22\textwidth,keepaspectratio}{
    \includegraphics{#1}
  }
}

\journal{Journal of Cultural Heritage}

\begin{document}

\begin{frontmatter}


\title{Vision-Language Based Expert Reporting for Painting Authentication and Defect Detection} 

\cortext[cor1]{Corresponding Author}

\author[label1]{Eman Ouda}
\author[label1]{Mohammed Salah} 
\author[label3]{Arsenii O. Chulkov}
\author[label4]{Gianfranco Gargiulo}
 \author[label5]{Gian Luca Tartaglia}
 \author[label2]{Stefano Sfarra}
\author[label1,aric]{Yusra Abdulrahman \corref{cor1}}

\affiliation[label1]{organization={Khalifa University of Science and Technology, Department of Aerospace Engineering},
            city={Abu Dhabi},
            country={United Arab Emirates}}
\affiliation[label3]{organization={National Research Tomsk Polytechnic University},
            country={Russia}}
              \affiliation[label4]{organization={Academy of Fine Arts of Naples},
            city={Naples},
            country={Italy}}
               \affiliation[label5]{organization={Academy of Fine Arts of Sassari},
            city={Sassari},
            country={Italy}}
            \affiliation[label2]{organization={Department of Industrial and Information Engineering and Economics (DIIIE), University of L’Aquila},
            city={L'Aquila I-67100},
            country={Italy}}
     \affiliation[aric]{organization={Advanced Research and Innovation Center (ARIC), Khalifa University of Science \& Technology},
            city={Abu Dhabi},
            country={UAE}}

\begin{abstract}
Authenticity and condition assessment are central to conservation decision-making, yet interpretation and reporting of thermographic output remain largely bespoke and expert-dependent, complicating comparison across collections and limiting systematic integration into conservation documentation. Pulsed Active Infrared Thermography (AIRT) is sensitive to subsurface features such as material heterogeneity, voids, and past interventions; however, its broader adoption is constrained by artifact misinterpretation, inter-laboratory variability, and the absence of standardized, explainable reporting frameworks. Although multi-modal thermographic processing techniques are established, their integration with structured natural-language interpretation has not been explored in cultural heritage.
A fully automated thermography–vision–language model (VLM) framework is presented. It combines multi-modal AIRT analysis with modality-aware textual reporting, without human intervention during inference. Thermal sequences are processed using Principal Component Thermography (PCT), Thermographic Signal Reconstruction (TSR), and Pulsed Phase Thermography (PPT), and the resulting anomaly masks are fused into a consensus segmentation that emphasizes regions supported by multiple thermal indicators while mitigating boundary artifacts. The fused evidence is provided to a VLM, which generates structured reports describing the location of the anomaly, thermal behavior, and plausible physical interpretations while explicitly acknowledging the uncertainty and diagnostic limitations.
Evaluation on two marquetries demonstrates consistent anomaly detection and stable structured interpretations, indicating reproducibility and generalizability across-samples. By translating multi-modal thermal evidence into standardized natural-language documentation, the proposed framework advances conservation thermography toward explainable, reproducible, and documentation-ready reporting grounded in active thermal imaging.

\end{abstract}



\begin{keyword}



Thermography \sep Painting Authentication \sep vision-language models \sep Non-Destructive Testing \sep Conservation Science.

\end{keyword}

\end{frontmatter}




\section{Introduction and Research Aim}
\label{sec: Introduction}

Cultural heritage objects are often complex, multi-layered structures in which each layer fulfills a specific structural or aesthetic role \cite{Yao2018, Gavrilov2014, Peeters2018}. Among these, paintings represent a particularly intricate category, typically comprising a support (e.g., canvas or wood panel), often prepared with a sizing layer to promote adhesion, followed by a ground layer, superimposed paint layers, and, in many cases, a protective varnish \cite{ Gavrilov2014}.
Although the historical and physical value of such objects is often associated with their support, the primary artistic significance is traditionally associated with the decorated surface \cite{Yao2018}. Assessing the condition and authenticity of objects of cultural heritage, particularly layered art, such as panel paintings, is therefore a multidimensional task that integrates material evidence, structural integrity, and historical context \cite{Gavrilov2014, Yao2018}.
 These assessments directly inform conservation planning, risk management, and market valuation, motivating the adoption of non-destructive testing (NDT) techniques capable of probing subsurface features without physical intervention \cite{Bendada2015, Liu2025}.

Among the available NDT approaches, Active Infrared Thermography (AIRT) has been successfully applied to paintings and painted panels to reveal structural defects, moisture-related features, and stratigraphic variations, providing information complementary to optical techniques such as infrared reflectography \cite{Bendada2015,Laureti2019,Vavilov29112024}. In AIRT, an external thermal stimulus is applied to the surface, and the resulting transient thermal response is recorded over time \cite{Liu2025, Alexakis2024, Peeters2018}. Subsurface discontinuities, such as air-filled delaminations or voids, locally disrupt heat diffusion and generate temperature contrasts that can be analyzed to infer internal structure \cite{Kunikata2025,Zhao2023}.
In layered paint systems, AIRT can reveal features such as gaps, delamination, retouchings, underdrawings, and moisture-related phenomena by capturing variations in thermal emission after excitation \cite{Yao2018,Ambrosini2010,Bendada2015,Kunikata2025,Rippa2021,Bison29112024}.
Despite these demonstrated capabilities, their routine use in conservation practice remains inconsistent. Practical limitations include heat-loss artifacts, uneven excitation, variability in post-processing methods, and the absence of standardized reporting protocols \cite{Liu2025,Cheng2022,Mishra2024}. As a result, interpretation often remains expert-dependent, reducing reproducibility and limiting systematic integration into conservation documentation.

Three practical factors limit the reproducibility and interpretability of thermographic analysis. First, boundary heat loss, non-uniform excitation, and environmental reflections introduce artifacts, particularly near edges, that can bias the detection of defects \cite{Cheng2022,Liu2025}. Second, although advanced processing techniques such as Principal Component Thermography (PCT), Thermographic Signal Reconstruction (TSR), and Pulsed Phase Thermography (PPT) have been developed to enhance feature extraction and reduce noise, their implementation depends on parameter selection and feature interpretation that vary across laboratories and devices, leading to inconsistent results even for similar objects.
Third, even when artifacts are mitigated, variability in parameter selection and interpretation of features between laboratories introduces inconsistencies in the reported results \cite{Liu2025,Mishra2024}.
To address these challenges, post-processing methods such as PCT, TSR, and PPT have been developed, each emphasizing complementary aspects of the thermal response. However, no single modality is sufficient in isolation, and interpretation remains dependent on expert judgment.


To date, integration of multi-modal AIRT with standardized, interpretable reporting has not been formalized in cultural heritage diagnostics. Although complementary thermographic modalities such as PCT, TSR, and PPT can enhance the visibility of features and reduce artifacts, their outputs remain numerical or image-based representations that require expert interpretation. 
Consequently, thermographic output remains primarily image-based representations that require expert interpretation, limiting their integration into standardized conservation workflows. Instead of relying solely on quantified thermal metrics interpreted manually by experts, vision–language models (VLM)s offer a promising paradigm to generate structured, human-interpretable reports from complex visual data \cite{iLi2025,Xuyan2025}. However, the unconstrained use of such models raises concerns about hallucination, semantic misalignment, and lack of physical grounding.

\subsection{Research Aim}
Motivated by the above, the aim of this paper is to introduce a framework in which a VLM performs modality-aware natural-language reasoning grounded in multi-modal thermal evidence. In particular, the paper addresses the aforementioned research gap by proposing an integrated framework that couples multi-modal Pulsed AIRT with consensus anomaly detection and structured VLM-assisted reporting. The VLM synthesizes complementary representations, articulates physically plausible explanations, explicitly acknowledges uncertainty, and generates structured documentation suitable for conservation practice.
The main contributions of this work are as follows:

\begin{itemize}
\item A structured VLM-based reporting framework is introduced to enable a modality-aware natural-language interpretation of thermographic evidence. The framework generates standardized interpretive reports that explicitly distinguish intentional artistic features from potential damage or later interventions and provide conservation-oriented guidance with transparent uncertainty statements.
\item A consensus anomaly detection strategy is developed that prioritizes cross-modality agreement and suppresses boundary-related and heat-loss artifacts, thereby improving the spatial focus and robustness of subsurface anomaly localization.
\item Experimental validation is provided through two case studies with differing thermal contrast characteristics, demonstrating cross-sample consistency, interpretability of structured reports, and reproducibility of anomaly detection under conservation-relevant conditions.
\end{itemize}



\subsection{Related Work} 

\label{sec: Relatedwork}

Cultural heritage objects, including books, panel and canvas paintings, sculptures, and archeological artifacts, are complex multilayered systems whose material stratigraphy reflects both artistic techniques and subsequent aging or restoration processes.
In paintings in particular, the superposition of support, ground, paint layers, varnish, and possible overpaints or restorations creates heterogeneous structures where subsurface defects may remain visually undetectable. These defects commonly include delamination, cracks, voids, air pockets, concealed underdrawings, and hidden signatures \cite{Gavrilov2014}.
Because such features are often embedded beneath surface layers, AIRT has become essential tools in conservation diagnostics and authentication studies. AIRT is a non-contact full-field NDT technique that detects subsurface defects by monitoring heat diffusion–induced surface temperature variations \cite{Zhao2023, Rippa2023}. Applications of AIRT in cultural heritage range from the evaluation of collagen-based materials, such as parchment, to the detection of concealed joining materials under decorative metal layers \cite{Mercuri2011}. In painting conservation, AIRT complements infrared reflectography by simultaneously revealing underdrawings and structural defects such as detachments and stratigraphic irregularities \cite{Bendada2015,Laureti2019}. However, the diagnostic performance of AIRT is strongly dependent on excitation uniformity, thermal property contrast, defect geometry, and acquisition conditions \cite{Meola2007}.

To enhance interpretability, several signal processing techniques have been developed. 
Time-domain approaches such as TSR reduce noise and improve defect contrast through polynomial fitting of logarithmic temperature decay curves. 
Frequency-domain methods, notably PPT, transform thermal signals via Fourier analysis, extracting phase information that is less sensitive to non-uniform heating and emissivity variations \cite{Zhao2023,Kunikata2025, salah_pca}. 
PCT, based on PCA and implemented by singular value decomposition, projects thermal sequences onto orthogonal components to suppress noise and highlight dominant thermal patterns \cite{Zhang2017}. 
More recently, deep learning approaches have been proposed to automate thermographic defect detection in artworks, including spatiotemporal convolutional models and semantic segmentation techniques that mitigate pigment-related emissivity effects \cite{Alexakis2024, salah_pt}. Despite these advances, practical limitations remain significant. Thermographic measurements are highly sensitive to non-uniform heating, boundary heat los. In complex and heterogeneous artworks, different processing modalities may emphasize different physical phenomena, leading to ambiguity or artifact-driven interpretation. Furthermore, most analyzes focus on single-modality representations, limiting cross-validation across complementary thermal descriptors. As a result, interpretation and reporting remain largely manual and difficult to reproduce in all studies \cite{Cheng2022,Liu2023}.

Parallel to advances in thermographic processing, VLMs have emerged as powerful multimodal architectures capable of jointly modeling visual and textual information \cite{Li2025,iLi2025}. By aligning image features with natural language representations, VLMs enable tasks such as visual question answering, semantic grounding, and automated report generation \cite{Xuyan2025,Li2024}. However, applying VLMs in specialized scientific domains presents notable challenges. These models are prone to hallucinations, multimodal misalignment, and limited grounding in domain-specific physical principles. In cultural heritage applications, data scarcity is a major bottleneck: high-quality image–text pairs are rare, expert annotations are costly, and acquisition artifacts further complicate training \cite{Li2025,Chen2025}. Although transfer learning, self-supervised learning, and few-shot adaptation strategies partially mitigate data limitations, concerns regarding interpretability, trustworthiness, and reproducibility remain \cite{Liu2025,Xuyan2025}.

Importantly, thermographic diagnostics and vision–language reasoning have largely evolved as independent research streams. IRT provides physics-based evidence of subsurface structure, yet translating multi-representational thermal data into standardized, reproducible expert narratives remains manual and expert-dependent \cite{Cheng2022}. In contrast, VLMs excel at structured narrative synthesis, but lack intrinsic physical grounding when applied without constraint \cite{iLi2025,Xuyan2025}. To date, a formalized framework that systematically links multi-modal thermographic representations (e.g., raw sequences, PCT, TSR, PPT outputs) to domain-constrained, standardized VLM-based reporting in paintings conservation has not been established. The proposed framework addresses this gap by integrating consensus-based multi-modal thermographic analysis with schema-constrained VLM reporting. By grounding language generation in physically interpretable thermal descriptors and enforcing structured reporting templates, the approach improves reproducibility, reduces subjective variability, and provides a transparent pathway from thermal evidence to conservation-relevant interpretation. This integration offers a unified, explainable workflow that bridges subsurface diagnostics and standardized expert communication.

\section{Materials and Methods}

\label{sec: Methodology}

\subsection{Data Collection and Specimens}

\begin{figure}[htb!]
    \centering
    \begin{subfigure}[b]{0.45\textwidth}
        \centering
        \includegraphics[width=\textwidth]{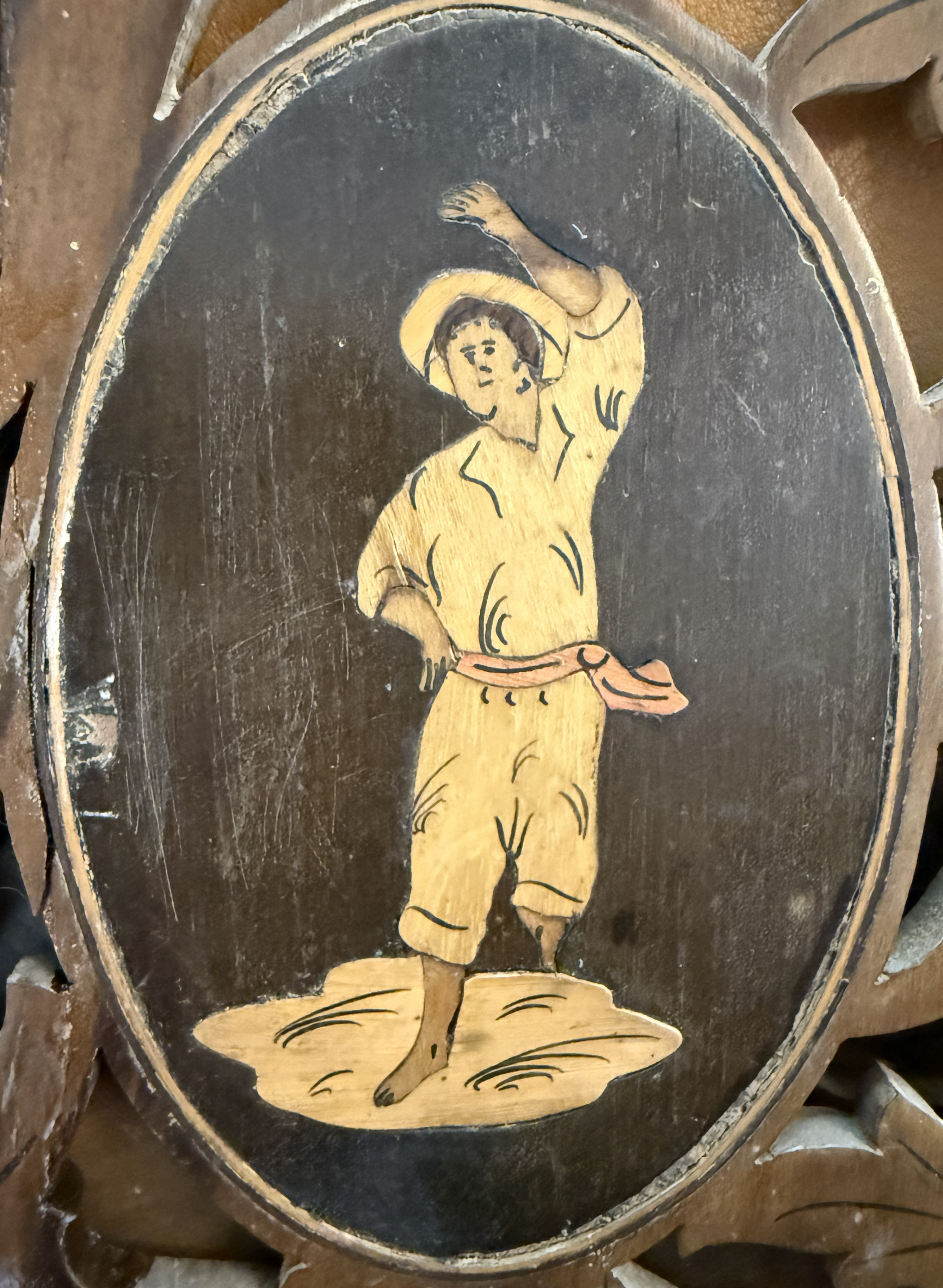}
        \caption{}
        \label{fig:optical_samplesb}
    \end{subfigure}
    \hfill
    \begin{subfigure}[b]{0.43\textwidth}
        \centering
        \includegraphics[width=\textwidth]{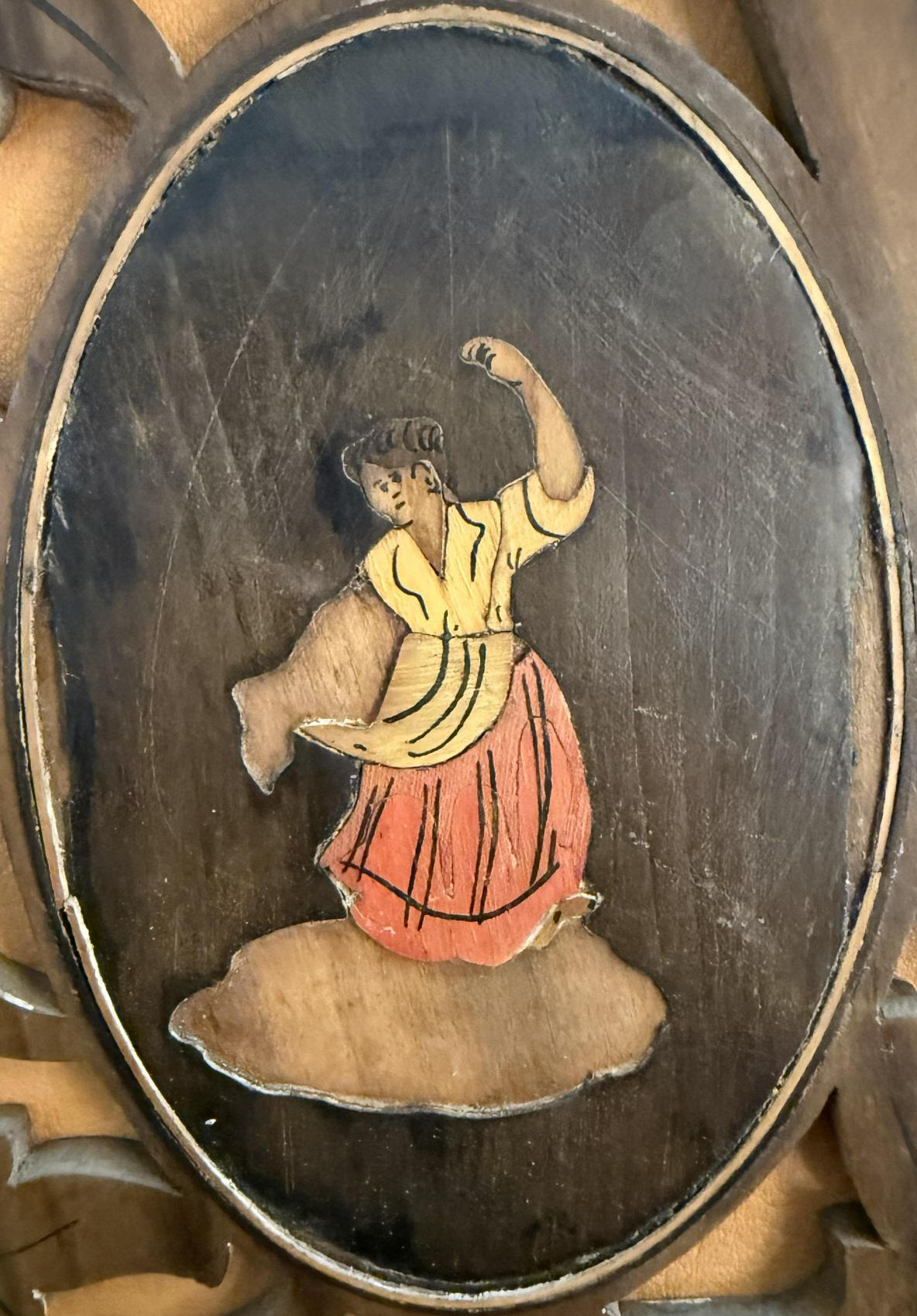}
        \caption{}
        \label{fig:optical_samplesg}
    \end{subfigure}
    \caption{
    Optical images of the Italian marquetry samples
analyzed in this study: (a) the ``Boy'' and (b) the ``Girl'' panel.}
    \label{fig:main}
\end{figure}

This study was conducted on two nineteenth-century Italian marquetry artworks, hereafter referred to as the ``Boy'' and ``Girl'' samples. The panels were selected due to the documented presence of both surface and subsurface defects associated with natural aging, environmental fluctuations, and multiple historical restoration interventions \cite{Chulkov2021}. Optical documentation of the two specimens is presented in Fig.~\ref{fig:main}, illustrating their conservation state before thermographic examination and providing visual reference for the distribution of visible deterioration phenomena.
The marquetry panels are made up of fruitwood veneers with an average thickness of approximately 0.5~mm, assembled according to the traditional Italian \textit{a buio} technique. In this method, individually cut decorative tesserae are adhered to a solid planar fruitwood support, which remains partially visible around the decorative composition. The material compatibility between the veneer and the support reduces the intrinsic differential hygroscopic response; however, the thin veneer layer makes the system particularly vulnerable to shear stresses and adhesive degradation at the veneer–support interface.

The original adhesive is presumed to be protein-based, consistent with workshops of the nineteenth-century. Nevertheless, both artworks underwent at least two restoration campaigns during the twentieth century. These interventions introduced non-original adhesives and locally incompatible restoration materials, generating mechanical and thermal heterogeneities within the stratigraphy. In the ``Boy'' sample, these additions are visually identifiable as dark restoration grout applied directly onto the support. Despite such modifications, the original patina remains largely preserved, and fine decorative details executed in Indian ink are still visible beneath the varnish layer, indicating limited abrasion of the pictorial surface.

From a conservation point of view, the two panels exhibit distinct but structurally comparable deterioration patterns.
The ``Boy'' sample (Fig.~\ref{fig:optical_samplesb}) is characterized by widespread detachments and localized deformation of the tesserae, predominantly in the background and arm regions. These detachments appear as partial lifting and loss of adhesion at the veneer–support interface, suggesting progressive weakening of the original adhesive layer. A localized lift is visible below the terminal section of the belt, indicating incomplete detachment with preserved peripheral bonding. The background areas show pronounced dirt encrustations, particularly in the upper and lower sections, while the remaining surface is covered by a relatively homogeneous layer of accumulated particulate matter. Localized greasy residues are also present, likely associated with previous handling or restoration treatments. The combined presence of detachments, surface contamination, and restoration grout indicates a complex stratigraphic condition that involves superficial and deeper structural alterations.
The ``Girl'' sample (Fig.~\ref{fig:optical_samplesg}) presents multiple areas of loss of tesserae that affect the regions of the feet, ground, and left arm. In these areas, the underlying wooden support is exposed, revealing heavy dirt encrustation and visible residues of the original protein adhesive.



\begin{figure}[htb!]
    \centering
\includegraphics[width=0.9\linewidth]{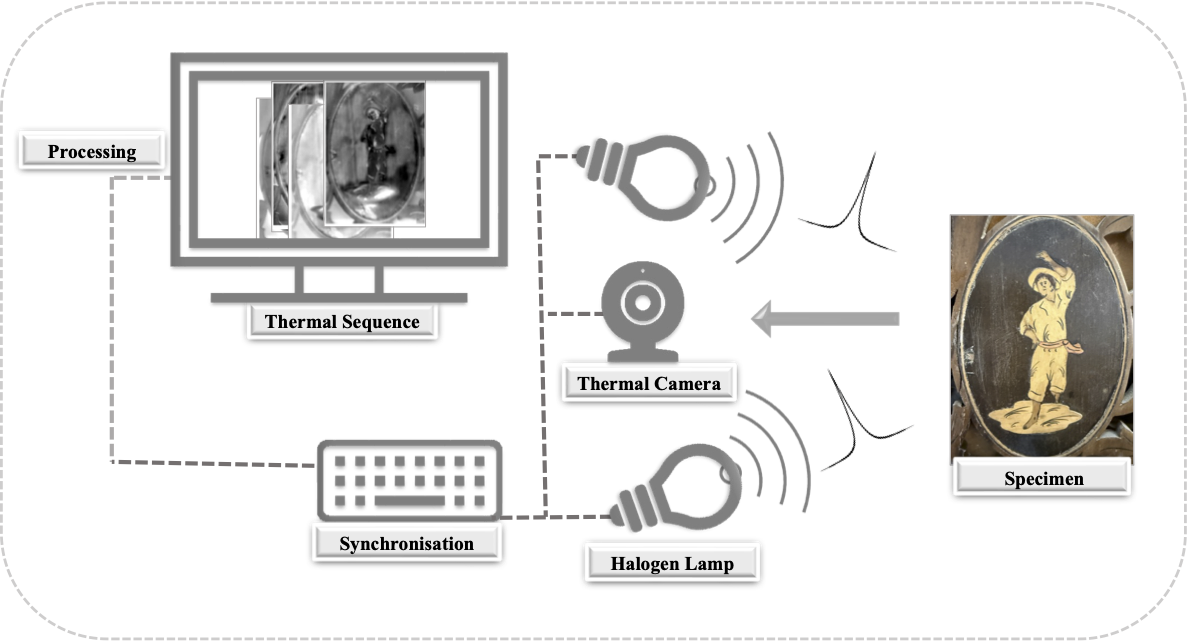}
    \caption{Active pulsed infrared thermography setup used for the acquisition of thermal sequences under conservation-safe conditions.}
    \label{fig:thermo_setup}
\end{figure}

The experimental configuration is shown in
Fig.~\ref{fig:thermo_setup}. Thermal excitation was provided by a Xenon flash tube that delivers an optical
energy of 3.2~kJ. The flash was applied in reflection mode at a distance of
approximately 1~m to promote uniform surface heating under conservation-safe
conditions. The thermal response was recorded using an FLIR Phoenix infrared
imager operating in the mid-wave infrared (MWIR) spectral range
(3--5~$\mu$m), which is well suited to detect small temperature variations
on painted and wooden surfaces.

Following flash excitation, sequences of thermograms were acquired during
the cooling phase to capture the full transient thermal response of each
sample. These temperature–time sequences form the basis for subsequent
thermographic processing, including PCT, TSR, and PPT analyses. The
acquisition protocol was designed to ensure repeatability and minimize
environmental influences, allowing a consistent comparison between the
``Boy'' and ``Girl'' marquetry specimens.

\subsection{Methodology}

\begin{figure}
    \centering    
    \includegraphics[width=\linewidth]{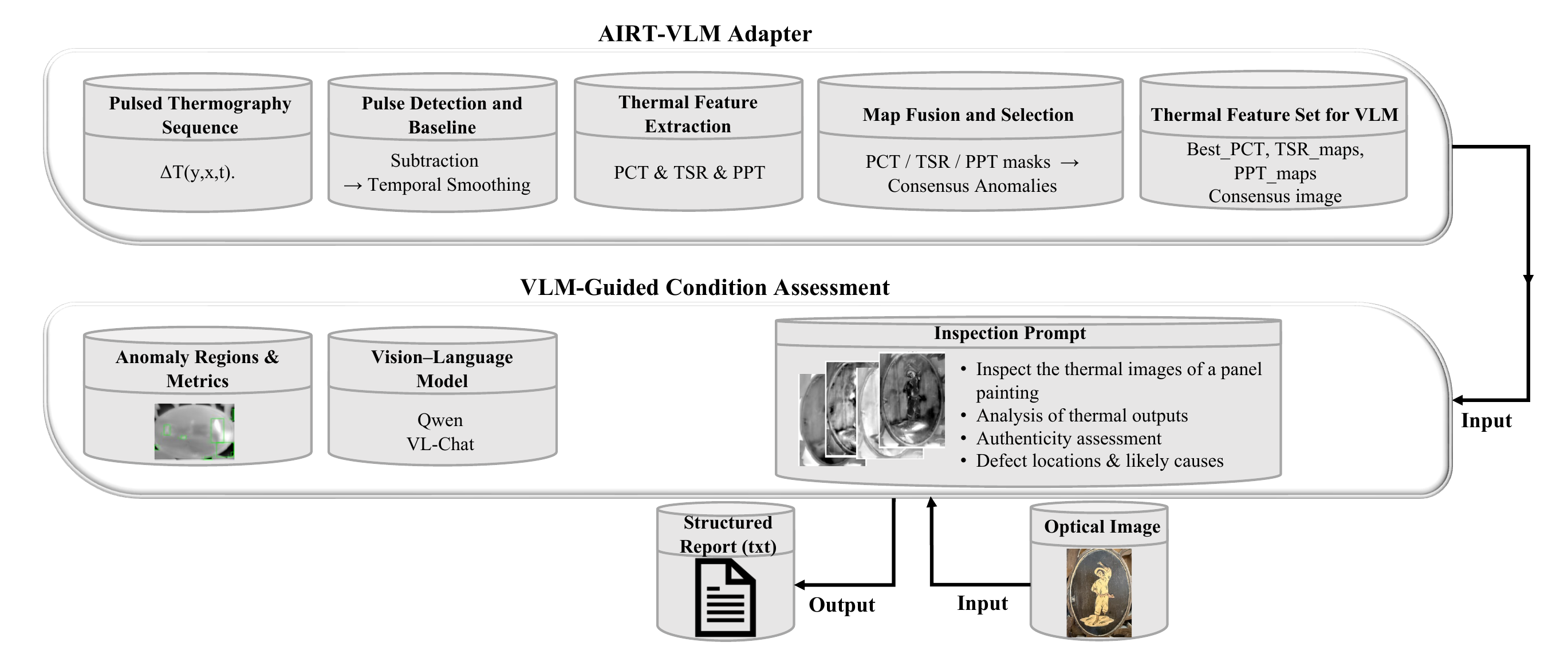}
    \caption{Thermography–VLM framework. Multi-modal thermographic maps are fused into a consensus anomaly map and provided, with the optical image, to a VLM that generates structured defect reports and quantitative summaries.}
    \label{fig: overallframework}
\end{figure}

This work proposes an integrated thermography–VLM framework for detecting, localizing, and interpreting subsurface anomalies in panel paintings (Fig.~\ref{fig: overallframework}), combining thermographic analysis with structured interpretive reporting. The workflow begins with conservation-safe acquisition of AIRT, automated pulse characterization, and selection of the region of interest (ROI) to ensure consistent preprocessing and baseline correction. The resulting thermal sequence is analyzed through complementary transforms, PCT to highlight heterogeneous temporal behavior, TSR to characterize cooling dynamics, and PPT to extract phase and amplitude features robust to non-uniform heating, after which anomaly maps from each modality are fused into a consensus representation that suppresses artifacts while preserving spatially coherent regions supported by multiple physical indicators. These fused thermographic outputs are subsequently projected onto the optical image for spatial context and provided to a VLM guided by a structured, conservation-oriented prompt, enabling the generation of standardized, modality-aware reports that explicitly preserve uncertainty and distinguish potential damage from intentional artistic features, thus establishing a reproducible and interpretable workflow for cultural heritage diagnostics.

\subsubsection{Thermographic Data Representation}

The objective of the thermographic preprocessing stage is to transform the
acquired pulsed sequence into a set of complementary physical representations (PCT, TSR, PPT) that jointly
characterize temporal variance, cooling dynamics, and frequency-domain
behavior. These representations form the quantitative foundation for anomaly
detection and are subsequently provided as structured inputs to the
VLM for high-level interpretive synthesis.

The analysis begins with the acquisition of a pulsed thermographic sequence,
which can be represented as a three-dimensional temperature field
\[
\Delta T(x,y,t),
\]
where $(x,y)$ denote spatial coordinates and $t$ represents discrete time.
For computational convenience, the sequence is treated as a temperature
cube; however, throughout this work, it is conceptually interpreted as a
time-resolved thermographic sequence. To restrict the analysis to the
painted surface and exclude mounting or background regions, a fixed ROI is selected. All subsequent processing is performed
exclusively within this ROI.

Before feature extraction, the thermographic sequence is preprocessed to
stabilize temporal behavior and ensure spatial consistency. Non-finite
values are replaced using the temporal median at each pixel. The onset of
the thermal pulse ($t_0$) and the maximum thermal response
time ($t_{\text{peak}}$) are automatically estimated from a smoothed 
spatially averaged temperature–time curve. These reference points enable
a consistent baseline correction across the ROI.

Following pulse characterization, the mean pre-pulse temperature at each
pixel is subtracted to obtain a baseline-corrected sequence. Optional
Savitzky--Golay filtering is then applied independently to each pixel
time series to suppress temporal noise while preserving the underlying
thermal dynamics. The resulting sequence serves as a stable input for
multi-modal thermographic processing.

For each derived thermographic representation (PCT magnitude, TSR slope,
PPT amplitude, and PPT phase-gradient), map intensities are subsequently
standardized over the ROI according to
\[
\tilde{M}(x,y) = \frac{M(x,y) - \mu_M}{\sigma_M},
\]
where $\mu_M$ and $\sigma_M$ denote the mean and standard deviation of
$M$ within the ROI. Thresholds are therefore applied in $z$-score units,
ensuring that anomaly detection criteria remain comparable across
sequences and between the ``Boy'' and ``Girl'' samples.

As a first representation, PCT \cite{RAJIC2002} decomposes the thermographic sequence into orthogonal spatial patterns corresponding to the dominant temporal modes. After reshaping the sequence into a matrix where each row corresponds to a pixel and each column corresponds to a time step, singular value decomposition is applied. The leading components form principal component maps that reduce temporal redundancy and enhance localized thermal heterogeneities potentially associated with subsurface defects.

To complement this variance-based analysis, TSR~\cite{tsr1,tsr2,tsr3} models post-pulse cooling behavior in the logarithmic domain. For each pixel, the temperature decay is approximated by a low-degree polynomial expressed in log–log space. From the fitted model, two scalar maps are extracted: a log-amplitude map and a slope map. The slope map represents the rate of change of the logarithmic temperature contrast with respect to logarithmic time and is particularly sensitive to variations in heat diffusion caused by subsurface structural discontinuities.

In addition to time-domain representations, PPT~\cite{ppt} analyzes the frequency content of the thermal response. A discrete Fourier transform is applied along the temporal dimension of the temperature signal. From this transformation, amplitude and phase maps are obtained at selected frequency bins. In this study, the amplitude map is used to identify regions of strong thermal contrast, while spatial gradients of the phase map are used to detect phase discontinuities (phase-edge response). The phase representation is particularly advantageous because it is less sensitive to non-uniform heating and surface emissivity variations.

Once these complementary thermographic maps are obtained, candidate anomaly regions are identified independently in each modality using thresholding strategies. High percentage thresholds are applied to the PCT magnitude and PPT amplitude maps, while standardized thresholds are applied to the TSR slope map. Phase-gradient maps are additionally thresholded to detect abrupt discontinuities. To suppress spurious detections, connected components below a minimum area threshold are removed.

Rather than interpreting each modality in isolation, a consensus anomaly mask is constructed by fusing the most complementary indicators, specifically, the TSR slope mask and the PPT phase, edge mask. This consensus representation retains regions consistently detected across modalities while reducing boundary-related artifacts, thereby providing a robust spatial representation of candidate subsurface features that is subsequently passed to the VLM for structured interpretation.

\subsubsection{VLM-Based Expert Reporting}

\begin{figure}[htb!]
    \centering
    \includegraphics[width=1\linewidth]{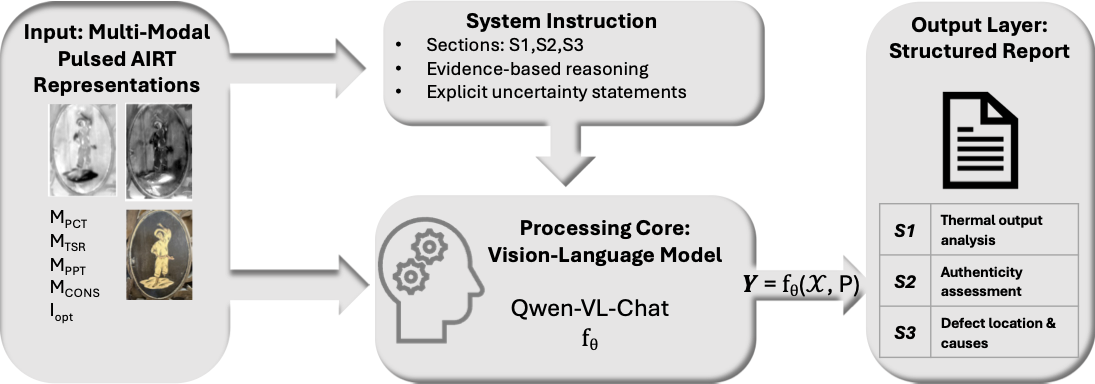}
    \caption{Proposed thermography–vision–language framework.}
    \label{fig:vlm_framework}
\end{figure}

As illustrated in Fig.~\ref{fig:vlm_framework}, the final stage of the pipeline
corresponds to the ``Processing Core: Vision–Language Model'' block, which
operates after thermographic detection and multi-modal fusion have been
completed.

With the spatial anomaly regions established (Detection → Fusion in
Fig.~\ref{fig:vlm_framework}), representative thermographic maps are selected
to support high-level interpretation. Let the thermographic representations be
defined as matrices over the ROI:
\[
\mathbf{M}_{\text{PCT}}, \quad
\mathbf{M}_{\text{TSR}}, \quad
\mathbf{M}_{\text{PPT}}, \quad
\mathbf{M}_{\text{CONS}},
\]
where each matrix encodes spatial information derived from a specific
thermographic modality (principal component map, TSR slope map,
PPT amplitude/phase-derived map and consensus anomaly mask, respectively.

A representative principal component is selected using a hybrid criterion
that combines the contrast-to-noise ratio and spatial overlap with the consensus
mask. This ensures that the retained component is both physically informative
and consistent with multi-modal anomaly detection.

As shown in the ``Input Layer: Multimodal Data'' block of
Fig.~\ref{fig:vlm_framework}, the final input to the VLM is defined as the
ordered set
\[
\mathcal{X} =
\left\{
\mathbf{M}_{\text{PCT}},
\mathbf{M}_{\text{TSR}},
\mathbf{M}_{\text{PPT}},
\mathbf{M}_{\text{CONS}},
\mathbf{I}_{\text{opt}}
\right\},
\]
where $\mathbf{I}_{\text{opt}}$ denotes the corresponding optical image
(provided when available), together with a structured conservation-oriented
prompt $P$.

At the encoding stage (``Processing Core'' in
Fig.~\ref{fig:vlm_framework}), a VLM with encoder parameters $\theta$
maps the multimodal input and prompt to a structured textual output:
\[
\mathcal{Y} = f_{\theta}(\mathcal{X}, P).
\]

Here, $f_{\theta}$ denotes a generic vision–language encoder–decoder model;
in this study, Qwen-VL-Chat~\cite{vlm} was employed, although the framework
is architecture-agnostic and can be implemented using an alternative VLM
architectures.

The prompt $P$ corresponds to the ``System Instruction'' block shown in
Fig.~\ref{fig:vlm_framework}. It is explicitly structured to constrain the
VLM outputs into a predefined reporting schema. Rather than allowing
free-form description, the model is instructed to generate a multi-section
report comprising:
\[
P = \{ S_1, S_2, S_3 \},
\]
where

\begin{itemize}
    \item $S_1$ (Thermal Output Analysis) requires a per-modality interpretation
 of PCT, PPT, TSR, and consensus maps, including spatial localization and
    plausible physical explanations;
    \item $S_2$ (Authenticity Assessment) requires cautious discussion of
    authenticity-related implications, with explicit separation between
    direct observations and hypotheses;
    \item $S_3$ (Defect Locations and Likely Causes) enforces a structured
 bulleted summary that specifies the location of the anomaly, the supporting modality, and
 a tentative physical interpretation.
\end{itemize}

In addition, system-level instruction constrains tone and epistemic
confidence by requiring evidence-based reasoning, explicit uncertainty
statements, and acknowledgment that thermography alone cannot establish
authenticity. The model is therefore not prompted to infer beyond the
available thermographic evidence, but to organize it within a
conservation-oriented interpretive framework.

Formally, the structured reporting process can be written as:
\[
\mathcal{Y} = f_{\theta}(\mathcal{X}, P_{\text{structured}}),
\]
where $P_{\text{structured}}$ encodes both the reporting sections and the
interpretive constraints.

This end-to-end progression,
\[
\text{Detection} \rightarrow \text{Fusion} \rightarrow \text{Encoding} \rightarrow \text{Structured Report},
\]
mirrors the layered architecture depicted in
Fig.~\ref{fig:vlm_framework} establishes a reproducible and explainable
pathway from thermal data to expert-oriented interpretation in cultural
heritage contexts.


\section{Results and Discussion}
\label{sec: Results}

The proposed framework was applied to AIRT sequences acquired from two marquetries, referred to as the ``Boy'' and ``Girl'' samples. 
For each dataset, an ROI was selected from the thermal cube, followed by automated pulse characterization, multi-modal thermographic analysis using PCT, TSR, and PPT, and consensus anomaly mapping. 
The results are presented together with their interpretation to highlight both the physical meaning of the thermographic outputs and the consistency of the framework across samples.

For both samples, the automated pulse detection procedure reliably identified pulse onset ($t_0$) and maximum thermal response time ($t_{\text{peak}}$), providing stable temporal reference points for baseline correction and post-pulse analysis. 
For the ``Girl'' marquetry, the pulse onset and peak response were detected at $t_0 = 73$ and $t_{\text{peak}} = 141$, respectively, while for the ``Boy'' marquetry, they occurred at $t_0 = 56$ and $t_{\text{peak}} = 124$. 
The base-median images exhibited comparable dynamic ranges for both datasets, with values spanning approximately $0.11$--$6.74$ for the ``Girl'' sample and $0.18$--$6.55$ for the ``Boy'' sample. 
These ranges indicate similar excitation levels and overall signal quality. 
Minor differences in pulse timing and baseline values were observed between the samples, but did not affect the stability of the preprocessing or the comparability of subsequent thermographic analyses.

\subsection{Thermographic Feature Maps and Consensus Analysis}

Representative thermographic feature maps obtained from PCT, TSR, PPT, and the consensus analysis derived are presented in Table~\ref{tab:thermal_outputs} for both the ``Boy'' and ``Girl'' marquetries. All images were extracted from the same ROI and contrast-normalized to allow consistent visual comparison across modalities.

\begin{table*}[htb!]
\centering
\caption{Representative thermographic outputs for the Boy and Girl marquetries,
including PCT (representative principal component), TSR slope maps, and PPT phase
maps, and the corresponding consensus anomaly maps. All images are derived
from the selected ROI and contrast-normalized for visualization.}
\label{tab:thermal_outputs}
\begin{tabular}{lcc}
\hline
Method & Boy sample & Girl sample \\
\hline
PCT  &
\includegraphics[width=0.45\textwidth]{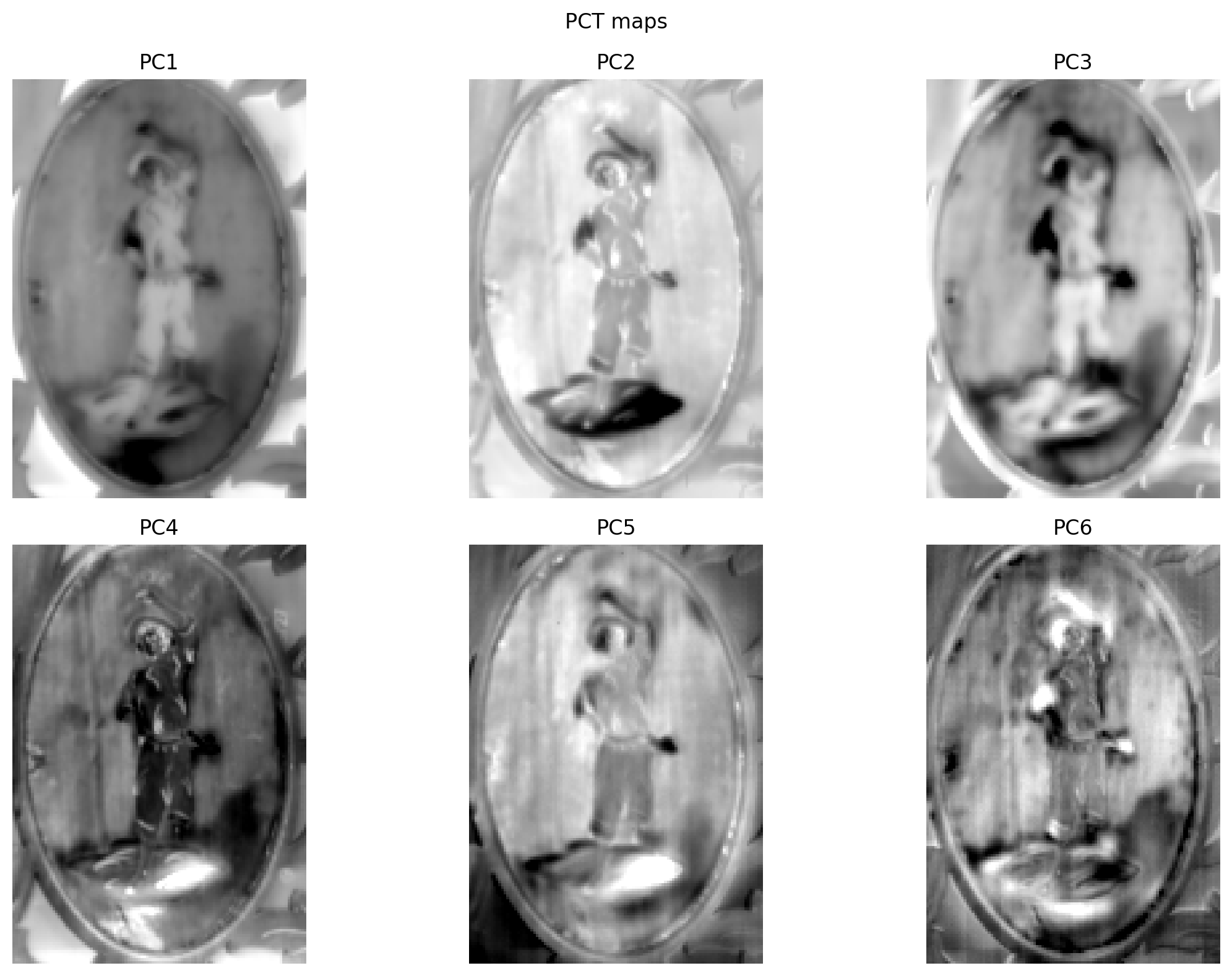} &
\includegraphics[width=0.45\textwidth]{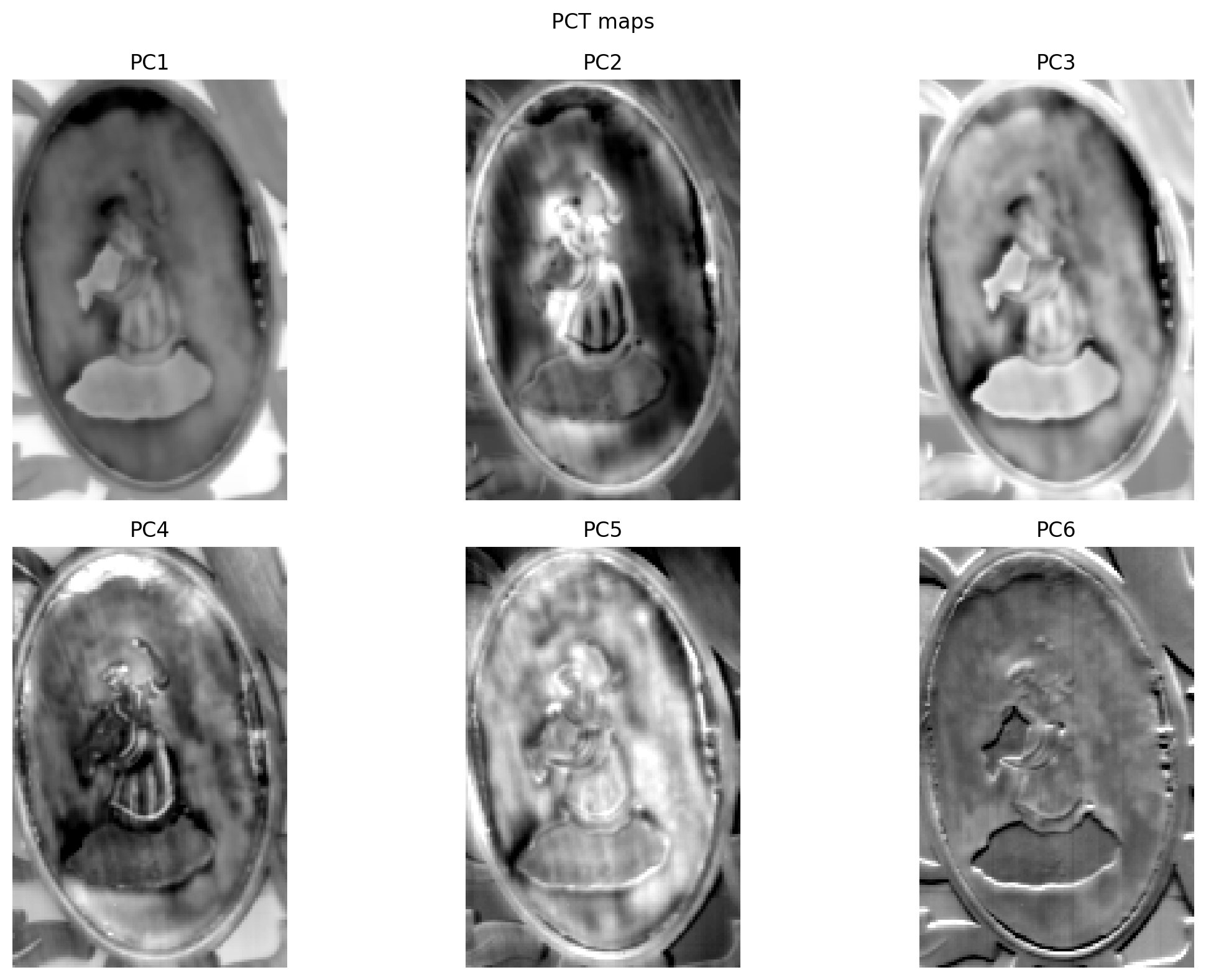} \\[6pt]

TSR  &
\includegraphics[width=0.45\textwidth]{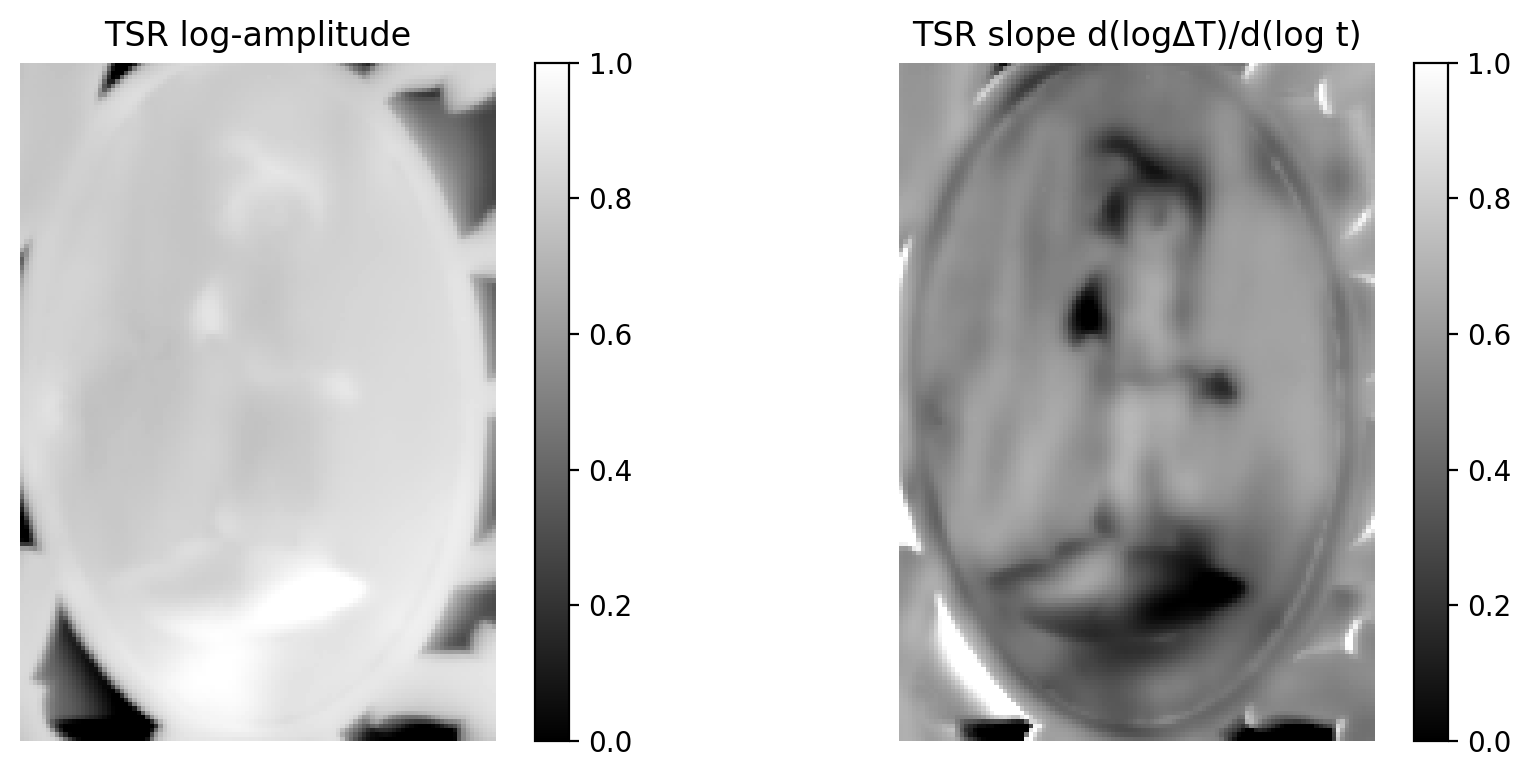} &
\includegraphics[width=0.45\textwidth]{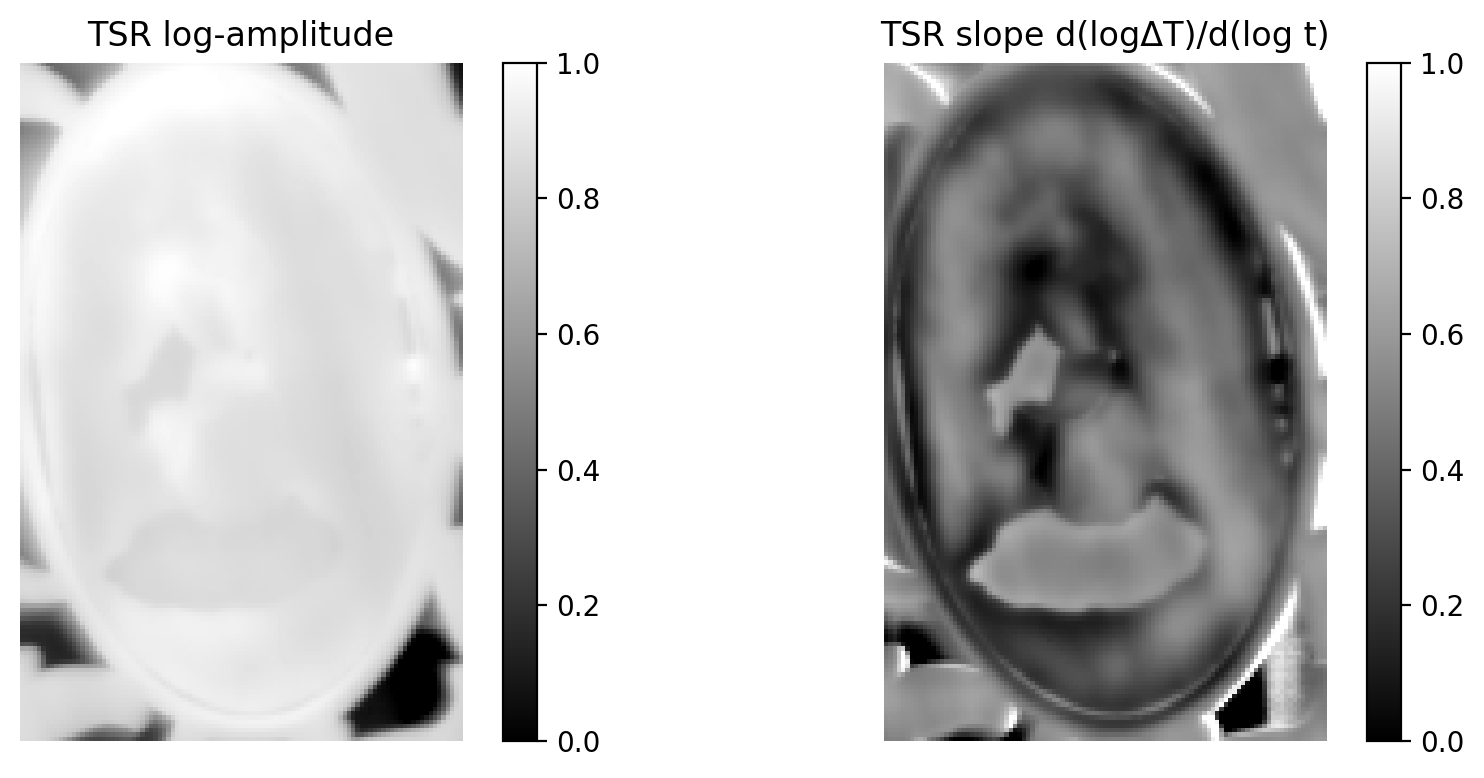} \\[10pt]

PPT  &
\includegraphics[width=0.45\textwidth]{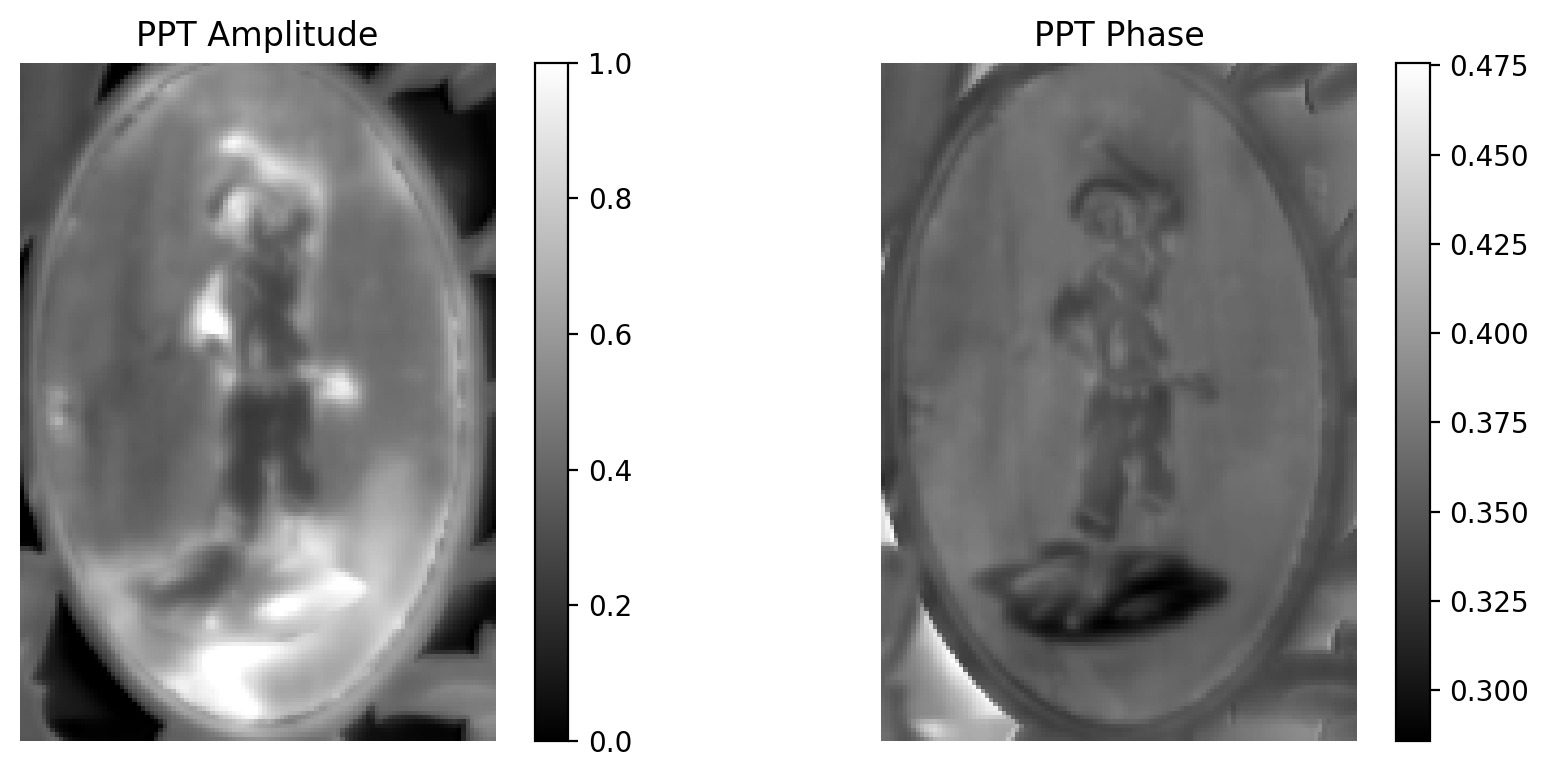} &
\includegraphics[width=0.45\textwidth]{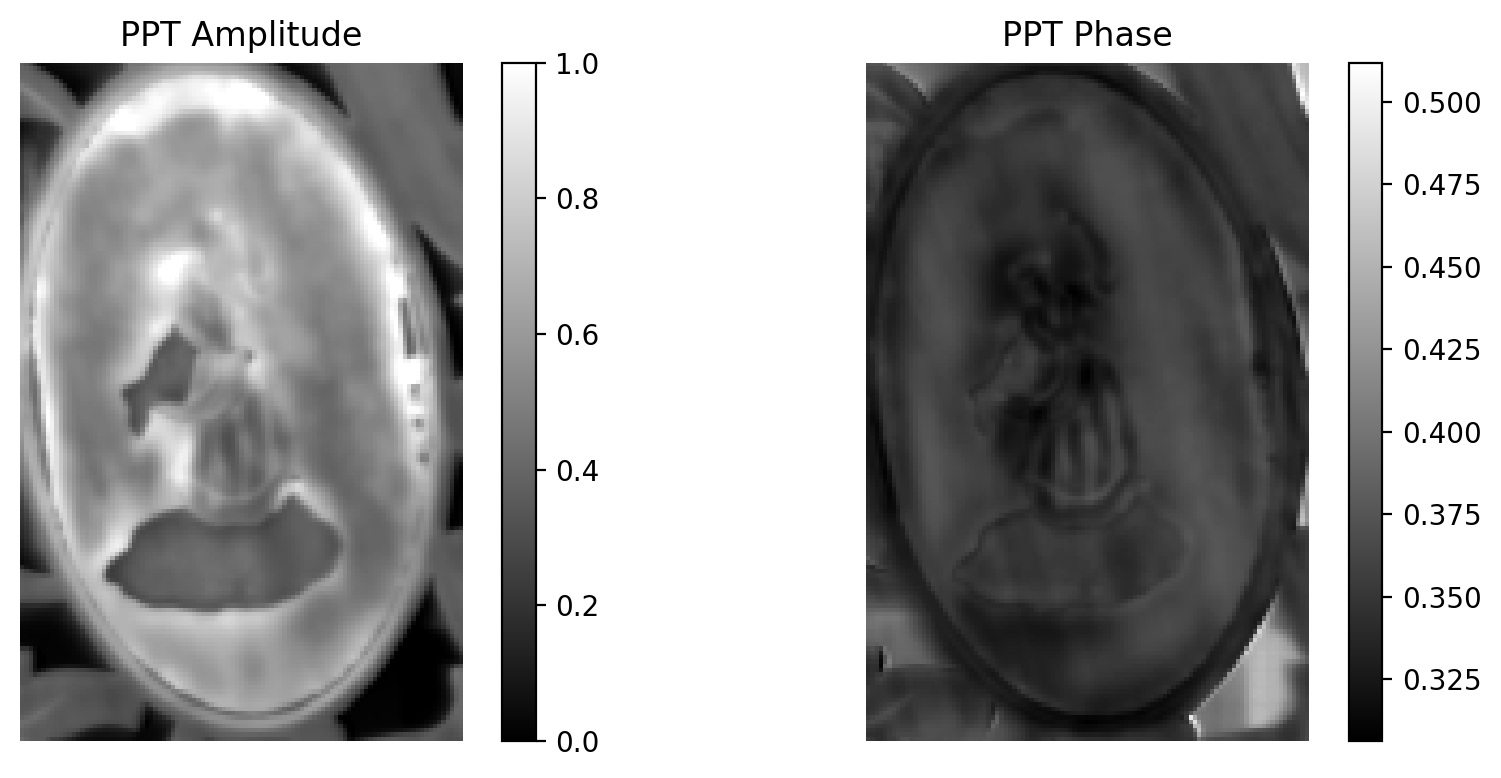} \\[6pt]

Consensus  &
\imgbox{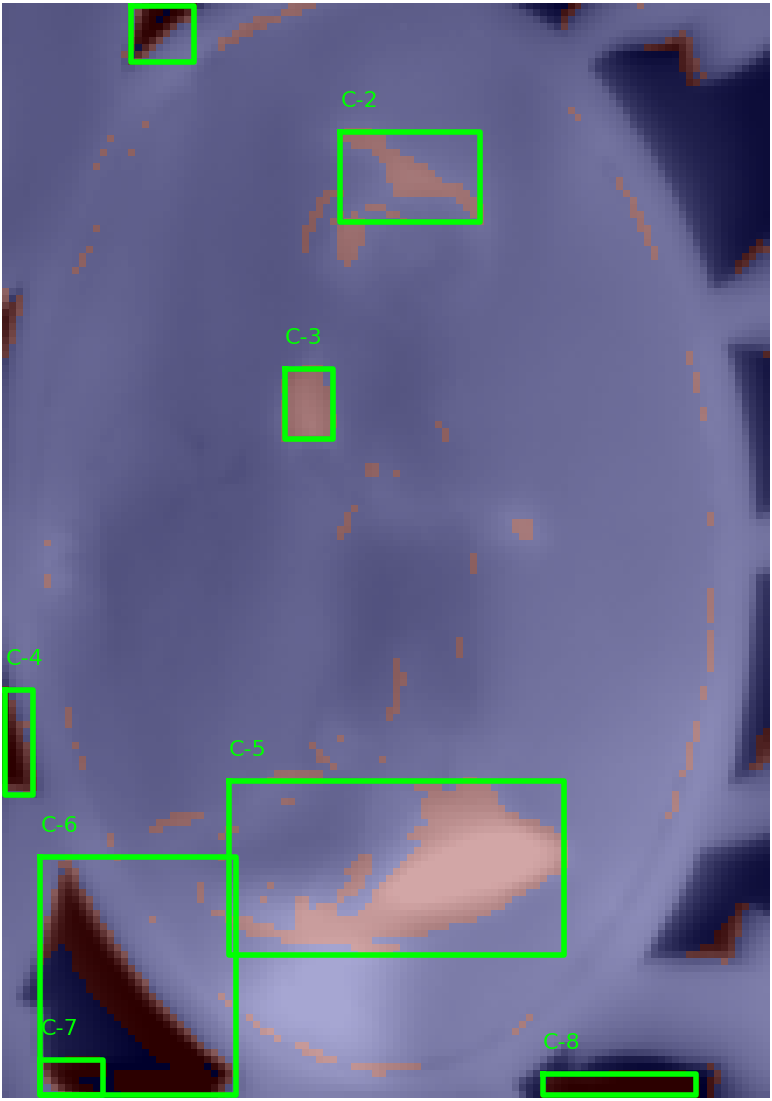} \imgbox{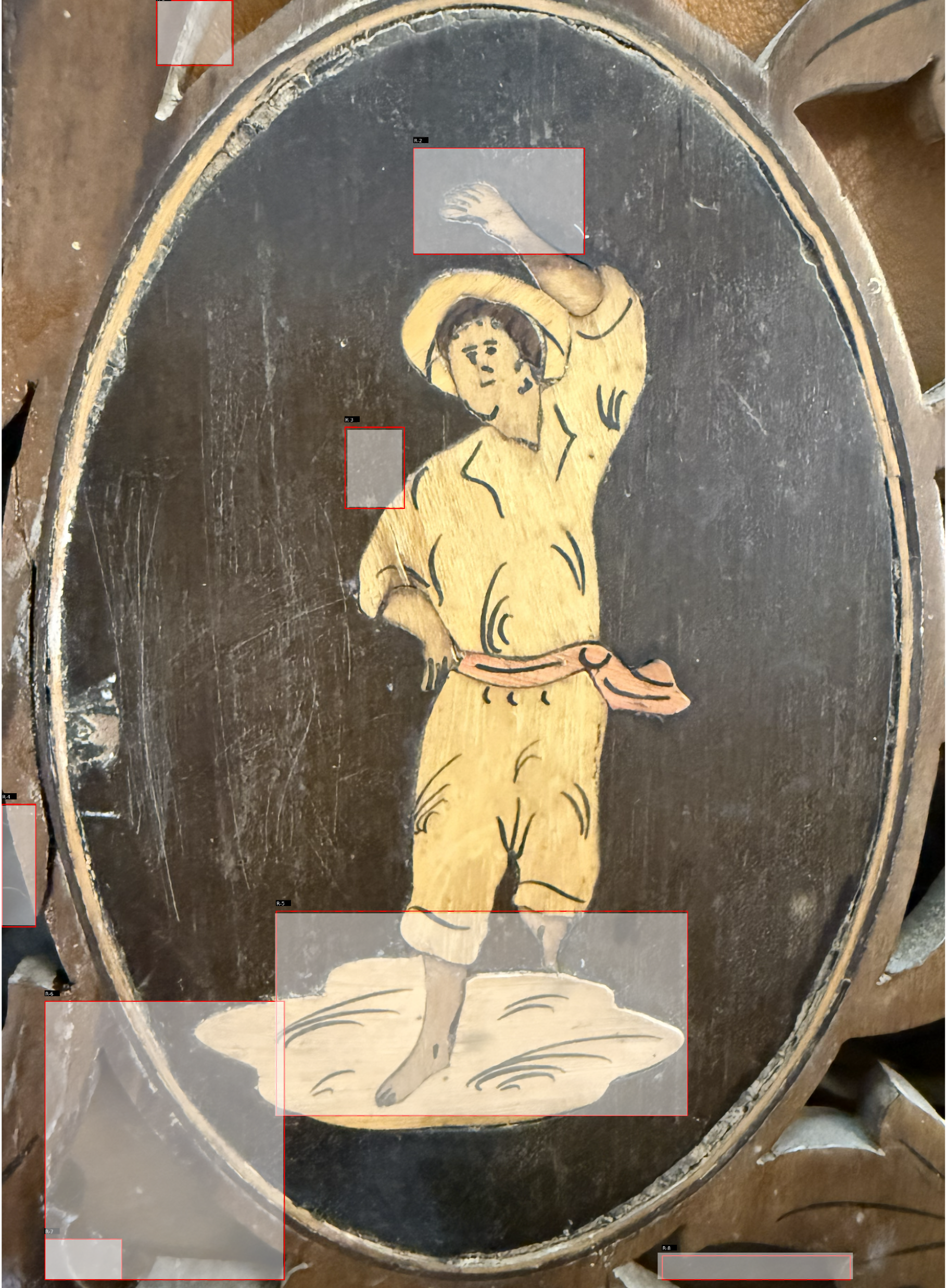} &
\imgbox{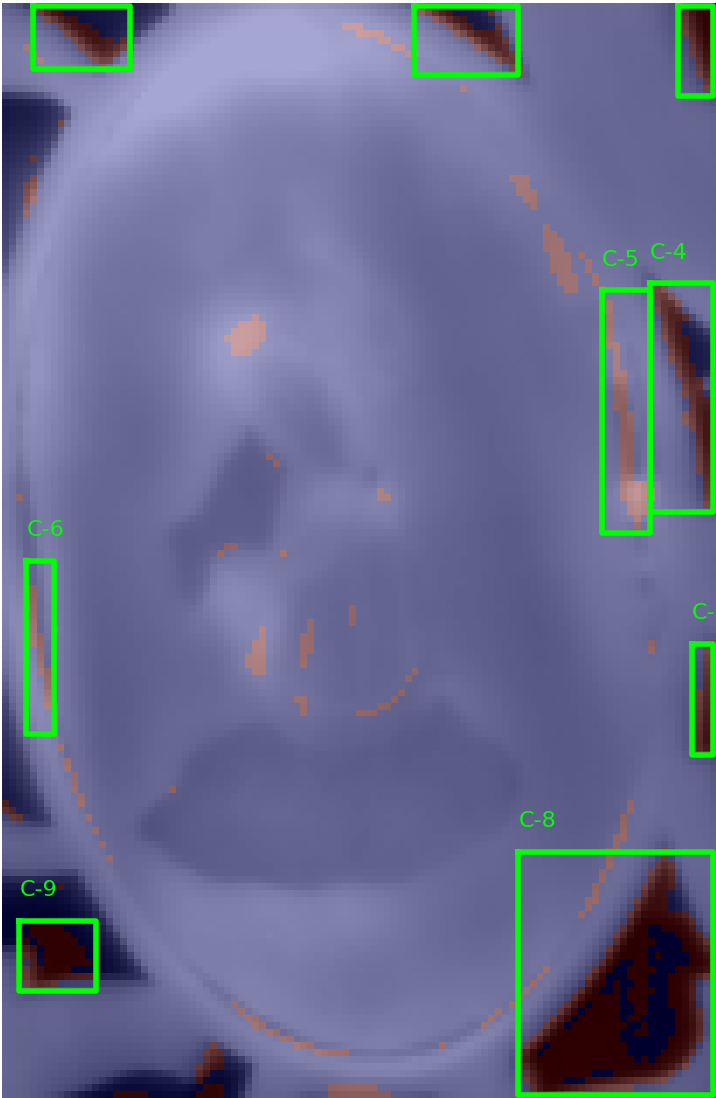} 
\imgbox{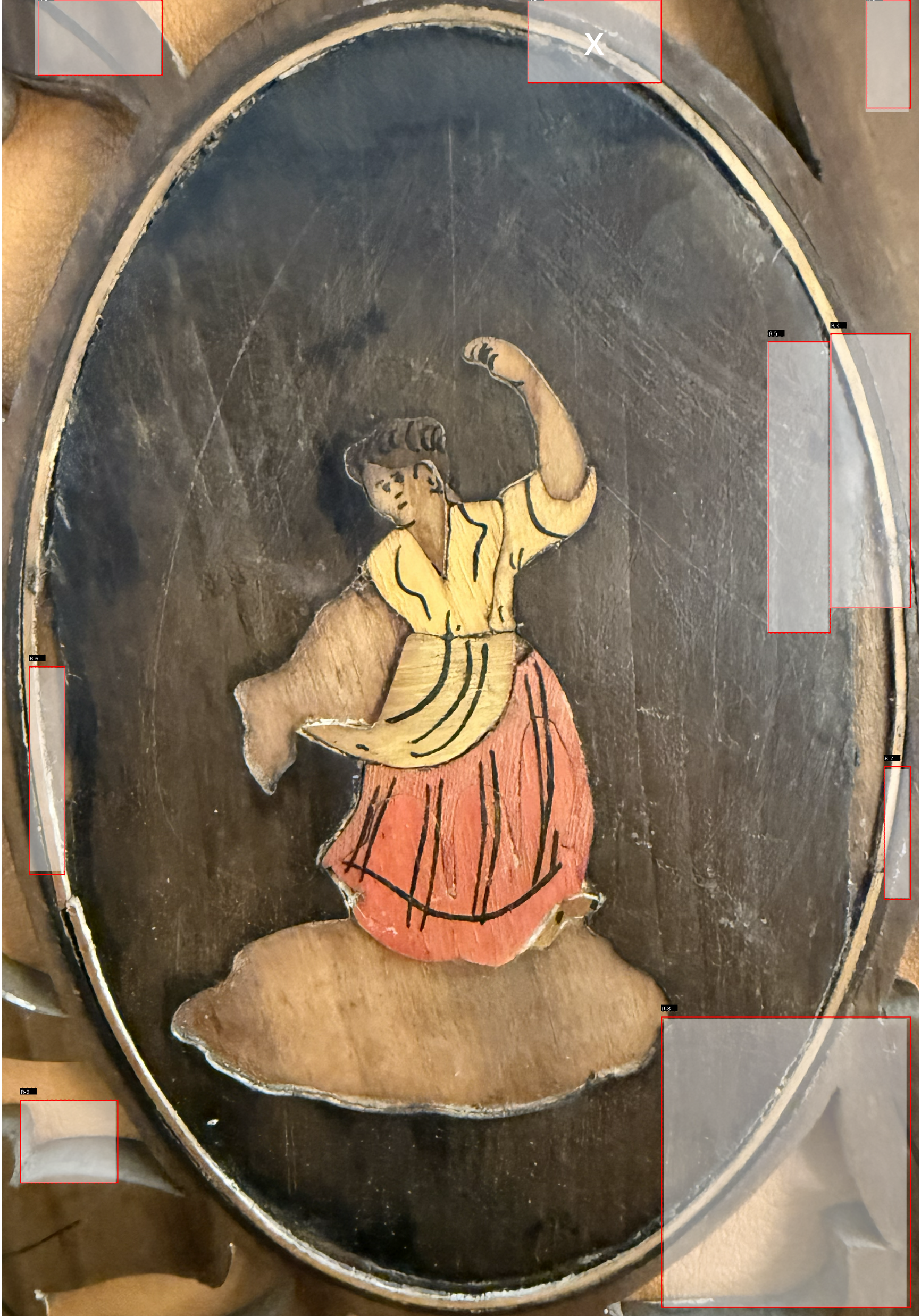} \\
\hline
\end{tabular}
\end{table*}





The PCT results show that the first principal component (PC1) in both samples primarily captures global thermal gradients and the geometry of the oval wooden support. Although PC1 preserves macroscopic structural information, it does not selectively isolate localized subsurface defects. In contrast, intermediate components (PC2--PC4) progressively suppress background diffusion trends and enhance spatially confined anomalies. For the ``Boy'' sample, these components clearly delineate the ground detachment, the shoulder anomaly, and the lower marquetry defect, although higher-order components introduce increasing noise. In the ``Girl'' sample, PCT reveals more spatially distributed irregularities, particularly in the bust region and along the frame perimeter, suggesting diffuse adhesion weakening rather than isolated large detachments.

The TSR outputs highlight the difference between log-amplitude and slope representations. Log-amplitude maps remain dominated by global heating and cooling gradients and exhibit limited sensitivity to localized discontinuities. In contrast, TSR slope maps enhance deviations from ideal diffusive behavior, improving defect visibility. In the ``Boy'' sample, slope imaging strengthens the contrast of the lower and shoulder detachments. In the ``Girl'' sample, slope maps reveal imperfections in the bust region and linear anomalies along the frame perimeter. These results confirm that derivative-based thermal parameters increase sensitivity to veneer–support interface disruptions in thin marquetry systems.

The PPT analysis further demonstrates the advantage of phase imaging over amplitude maps. Amplitude representations are influenced by surface emissivity variations, varnish heterogeneity, and curvature effects, which reduce their specificity for subsurface defects. In contrast, phase maps provide clearer and more spatially coherent representations of structural anomalies. For the ``Boy'' panel, phase imaging distinctly enhances the ground detachment, shoulder anomaly, and lower marquetry defect. For the ``Girl'' panel, phase contrast highlights adhesion weaknesses in the bust area, detachments along the frame, and regions associated with tesserae loss. The reduced sensitivity of phase to non-uniform heating and surface artifacts makes it particularly suitable for identifying veneer–support interface separations.

Across all individual modalities, no single technique consistently captures all anomalous regions. Each representation emphasizes different aspects of altered thermal behavior, underscoring the complementary nature of PCT, TSR slope, and PPT phase analysis.
The consensus anomaly maps shown in the final row of Table~\ref{tab:thermal_outputs} integrate the recurrent defect signatures detected across multiple thermographic indicators. By retaining only regions supported by more than one modality, this fusion strategy suppresses isolated noise responses, boundary-related heat-loss artifacts, and curvature-driven false positives visible in individual maps. In both samples, the resulting consensus maps exhibit a limited number of spatially coherent anomaly regions with improved interpretability and robustness. The ``Boy'' sample shows relatively concentrated and well-defined detachments, whereas the ``Girl'' sample presents a more fragmented pattern of distributed adhesion weaknesses.

\subsection{Anomaly Masks and Quantitative Comparison}
To quantify the spatial extent of the detected features, binary anomaly masks were derived from PCT magnitude, TSR slope, PPT amplitude, and PPT phase-edge maps using identical percentile-based and $z$-score thresholds for both samples. The relative spatial extent of each mask is summarized in Table~\ref{tab:mask_areas}, which reports the percentage of the ROI classified as anomalous by each modality, together with the final consensus anomaly area and the number of detected regions.

As summarized in Table~\ref{tab:mask_areas}, the PCT and PPT-amplitude masks remain highly sparse in both samples, each covering approximately 1\% of the ROI and highlighting only the most pronounced localized deviations in thermal behavior. In contrast, TSR-slope and PPT phase-edge masks occupy a larger fraction of the ROI, reflecting their sensitivity to broader variations in cooling rates and phase gradients.

\begin{table}[htb!]
  \centering
  \caption{Area of anomaly masks and consensus regions for the Boy and Girl samples.}
  \label{tab:mask_areas}
  \begin{tabular}{lcc}
    \hline
    Mask / Map             & Boy area (\%) & Girl area (\%) \\
    \hline
    PCT                    & 1.00         & 1.00          \\
    TSR slope              & 5.68         & 3.80          \\
    PPT amplitude          & 1.00         & 1.00          \\
    PPT phase-edge         & 5.00         & 5.00          \\
    \hline
    Consensus anomaly mask & 8.84 (8 regions) & 7.02 (9 regions) \\
    \hline
  \end{tabular}
\end{table}

The final consensus anomaly mask occupies 8.84\% of the ROI for the ``Boy'' sample and 7.02\% for the ``Girl'' sample, remaining below 10\% in both cases. The ``Boy'' marquetry yields eight consensus regions with a slightly larger total area, whereas the ``Girl'' marquetry yields nine regions with a smaller combined area. Under identical processing parameters, this suggests that the ``Boy'' panel contains fewer but more spatially extended anomalies, while the ``Girl'' panel exhibits a more fragmented pattern of localized features.


\subsection{Vision--Language Model Interpretation}

Thermographic output and consensus anomaly maps were analyzed using a VLM guided by a structured, conservation-oriented prompt. An illustrative example of the structured VLM output for the ``Boy'' marquetry is presented in Fig.~\ref{fig:vlm_report_example_boy}.
For clarity,
only the principal sections of the generated report are reproduced, including
the analysis of thermal outputs and the associated uncertainty statement. The
figure serves to exemplify the reporting structure rather than to reproduce
the full response of the model.

\begin{figure}[htb!]
\centering
\fbox{
\begin{minipage}{0.95\linewidth}
\small
\textbf{Illustrative Example of Structured VLM Output (Boy Sample)} \\[4pt]

\textbf{1. Analysis of Thermal Outputs} \\
\emph{PCT}: Mostly stable thermal distribution with localized anomalies,
particularly in the background and along the edges of the painting. \\
\emph{PPT}: Generally stable response with localized phase anomalies near the
edges and around the figure of the boy. \\
\emph{TSR}: Heterogeneous cooling behavior with regions of distinct decay
dynamics. \\
\emph{Consensus}: Several anomalies supported by multiple modalities,
including a prominent region near the center of the painting (corresponding
to the figure of the boy) and smaller anomalies along the edges and in the
background. \\[4pt]

\textbf{2. Authenticity Assessment (Cautious)} \\
The observed thermal behavior does not provide definitive evidence of
authenticity. Repeated anomalies along the left edge and localized phase
anomalies near the figure suggest possible restoration or fills; however,
further multi-modal confirmation is required. Anomalies in the background
and along the edges may reflect changes in composition or panel
heterogeneity, but these observations alone are insufficient to establish
authenticity. \\[4pt]

\textbf{3. Defect Locations and Likely Causes} \\
\emph{Location}: Lower left region, near the left edge, and along the edges
of the painting. \\
\emph{Likely causes}: Possible retouching or fills suggested by anomalies
observed in the PCT and PPT maps and reinforced by the consensus mask.
Localized phase anomalies near the figure may indicate past repairs or
material heterogeneity in the paint and ground layers. \\[4pt]

\end{minipage}
}
\caption{Illustrative example of a structured vision--language model (VLM)
report generated for the ``Boy'' marquetry.}
\label{fig:vlm_report_example_boy}
\end{figure}

For the ``Boy'' marquetry, the VLM described the overall thermal behavior as largely stable with localized anomalies supported by multiple modalities, particularly along panel edges, in background areas, and near the central figure. 
These regions were associated with plausible physical explanations such as material heterogeneity, previous interventions, or structural differences in the paint and ground layers, without asserting definitive material diagnoses or authenticity claims.
For the ``Girl'' marquetry, the VLM similarly identified coherent global thermal behavior with a limited number of localized anomalies, including a prominent region in the lower left area and additional features near edges and around the figure. 
As with the ``Boy'' sample, all interpretations were framed cautiously as hypothetical and contingent on confirmation through complementary imaging and material analysis.

Across both samples, the VLM outputs remained consistent with the structured prompting framework: per-modality observations were clearly separated from hypotheses, explanations were grounded in thermal transport principles, and authenticity-related considerations were presented in explicitly non-definitive terms. Despite differences in the painted content and spatial distribution of features, the framework exhibited consistent behavior across the ``Boy'' and ``Girl'' marquetries.
In both cases, multi-modal thermographic analysis identified localized anomalies, consensus fusion reduced artifact-driven detections, and quantitative metrics fell within comparable ranges. This consistency across independent samples supports the applicability of the proposed framework beyond a single case study and suggests that the integration of multi-modal thermography with structured VLM interpretation provides a generalizable approach for standardized analysis and reporting in cultural heritage thermography.

The consensus mechanism plays a central role in improving reliability. By enforcing cross-modality agreement and down-weighting edge-proximal responses, the framework mitigates common thermographic failure modes while preserving sensitivity to interior features, an essential consideration in panel paintings, where frames and heterogeneous stratigraphy can amplify boundary heat-loss effects. Collectively, the integration of PCT, TSR, and PPT within this consensus-based strategy produces physically grounded and spatially constrained anomaly maps across distinct painted panels. The consistent behavior observed in both the ``Boy'' and ``Girl'' marquetries demonstrates cross-sample stability and supports the broader applicability of the approach, providing a robust basis for structured VLM interpretation and standardized conservation reporting.

\section{Conclusion}
\label{sec: Conclusion}

This study introduced an integrated framework for AIRT in cultural heritage, combining thermographic processing with structured VLM reporting. 
The pipeline comprises conservation-safe acquisition, multi-modal feature extraction using PCT, TSR, and PPT, thermographic consensus anomaly detection to suppress boundary artifacts, and standardized textual reporting under a domain-constrained prompting strategy.
Applied to two marquetries, the framework produced spatially focused anomaly maps supported by multiple thermographic indicators and demonstrated consistent behavior across samples.
The consensus strategy improved robustness by reducing the reliance on any single modality, while the
VLM component generated structured, modality-aware summaries that preserved uncertainty and distinguished intentional features from potential damage.
Together, these elements establish a reproducible and interpretable workflow for thermographic analysis in conservation practice.

The proposed framework is the first to integrate multi-modal thermographic analysis with VLM–assisted reporting for cultural heritage applications, addressing the lack of standardized and explainable thermographic interpretation.
It supports conservation practice by combining spatially constrained anomaly detection with structured, transparent narratives for condition assessment, documentation, and cautious evidence summarization. 
Designed to augment expert judgment rather than replace it, the framework also provides a scalable template for integrating additional non-destructive techniques, with the VLM serving as a unified narrative interface across heterogeneous data sources.


Future work will extend testing to a wider range of artworks and incorporate complementary techniques such as X-ray fluorescence (XRF) and hyperspectral imaging.
The development of domain-adapted VLMs tailored to cultural heritage applications will also be investigated to improve terminological consistency and further support standardized and transparent reporting practices.


\section*{Acknowledgments}
This research was funded by Khalifa University of Science and Technology through the [Advancing Non-Destructive Testing (NDT) through Innovative Integration of Infrared Thermography (IRT) and Emerging Technologies in Aerospace Applications] under Project ID: KU-[INT]-[FSU]-[2024]-[8474000660].

  \bibliographystyle{elsarticle-num} 


\bibliography{ArtworkMendeley}

@article{Bendada2015,
  title={Subsurface imaging for panel paintings inspection: a comparative study of the ultraviolet, the visible, the infrared and the terahertz spectra},
  author={Bendada, A and Sfarra, S and Ibarra- Castanedo, C and Akhloufi, M and Caumes, J- P and Pradere, C and Batsale, J- C and Maldague, X},
  journal={Opto-Electronics Review},
  volume={23},
  number={1},
  pages={90--101},
  year={2015},
  publisher={De Gruyter Open}
}

@article{Cheng2022,
   author = {Liangliang Cheng and Zongfei Tong and Shejuan Xie and Mathias Kersemans},
   doi = {10.1016/j.compstruct.2022.115543},
   issn = {02638223},
   journal = {Composite Structures},
   month = {6},
   publisher = {Elsevier Ltd},
   title = {IRT-GAN: A generative adversarial network with a multi-headed fusion strategy for automated defect detection in composites using infrared thermography},
   volume = {290},
   year = {2022}
}

@misc{Zhao2023,
   author = {Xinfeng Zhao and Yangjing Zhao and Shunchang Hu and Hongyan Wang and Yuyan Zhang and Wuyi Ming},
   doi = {10.3390/s23218780},
   issn = {14248220},
   issue = {21},
   journal = {Sensors (Basel, Switzerland)},
   month = {10},
   pmid = {37960480},
   title = {Progress in Active Infrared Imaging for Defect Detection in the Renewable and Electronic Industries},
   volume = {23},
   year = {2023}
}

@article{Alexakis2024,
   author = {Emmanouil Alexakis and Ekaterini T. Delegou and Philip Mavrepis and Antonis Rifios and Dimosthenis Kyriazis and Antonia Moropoulou},
   doi = {10.1016/j.cscm.2024.e02889},
   issn = {22145095},
   journal = {Case Studies in Construction Materials},
   month = {7},
   publisher = {Elsevier Ltd},
   title = {A novel application of deep learning approach over IRT images for the automated detection of rising damp on historical masonries},
   volume = {20},
   year = {2024}
}

@article{Kunikata2025,
   author = {Saki Kunikata and Yuko Tsuchiya and Kaori Fukunaga},
   doi = {10.1016/j.culher.2025.04.008},
   issn = {12962074},
   journal = {Journal of Cultural Heritage},
   month = {5},
   pages = {295-304},
   publisher = {Elsevier Masson s.r.l.},
   title = {Delamination of wax-resin linings in oil paintings: Visualization and analysis using infrared active thermography and terahertz time-domain imaging},
   volume = {73},
   year = {2025}
}

@article{Liu2023,
   author = {Kaixin Liu and Kai Lun Huang and Stefano Sfarra and Jianguo Yang and Liu Yi and Yao Yuan},
   doi = {10.1080/17686733.2021.2019658},
   issn = {21167176},
   issue = {1},
   journal = {Quantitative InfraRed Thermography Journal},
   pages = {25-37},
   publisher = {Taylor and Francis Ltd.},
   title = {Factor analysis thermography for defect detection of panel paintings},
   volume = {20},
   year = {2023}
}

@article{Yao2018,
   author = {Yuan Yao and Stefano Sfarra and Susana Lagüela and Clemente Ibarra-Castanedo and Jin Yi Wu and Xavier P.V. Maldague and Dario Ambrosini},
   doi = {10.1016/j.ijthermalsci.2017.12.036},
   issn = {12900729},
   journal = {International Journal of Thermal Sciences},
   month = {4},
   pages = {143-151},
   publisher = {Elsevier Masson SAS},
   title = {Active thermography testing and data analysis for the state of conservation of panel paintings},
   volume = {126},
   year = {2018}
}

@misc{Liu2025,
   author = {Yi Liu and Yuan Yao and Fumin Wang and Stefano Sfarra and Kaixin Liu},
   doi = {10.1080/17686733.2025.2540662},
   issn = {21167176},
   journal = {Quantitative InfraRed Thermography Journal},
   publisher = {Taylor and Francis Ltd.},
   title = {Review of unsupervised machine learning methods in active infrared thermography for defect detection and analysis},
   year = {2025}
}

@misc{Mishra2024,
   author = {Mayank Mishra and Paulo B. Lourenço},
   doi = {10.1016/j.culher.2024.01.005},
   issn = {12962074},
   journal = {Journal of Cultural Heritage},
   month = {3},
   pages = {536-550},
   publisher = {Elsevier Masson s.r.l.},
   title = {Artificial intelligence-assisted visual inspection for cultural heritage: State-of-the-art review},
   volume = {66},
   year = {2024}
}

@article{Chen2025,
   author = {Baozhong Chen and Xiaonan Wang},
   doi = {10.1504/ijict.2025.10070954},
   issn = {1466-6642},
   issue = {13},
   journal = {International Journal of Information and Communication Technology},
   publisher = {Inderscience Publishers},
   title = {Artificial intelligence for cultural heritage: digital image processing-based techniques and research challenges},
   volume = {26},
   year = {2025}
}

@article{Meola2007,
   author = {Carosena Meola},
   doi = {10.1016/j.matlet.2006.04.120},
   issn = {0167577X},
   issue = {3},
   journal = {Materials Letters},
   month = {2},
   pages = {747-750},
   title = {A new approach for estimation of defects detection with infrared thermography},
   volume = {61},
   year = {2007}
}

@article{iLi2025,
   author = {Xiang Li and Yu Sun and Jia Lin and Like Li and Ting Feng and Shen Yin},
   doi = {10.23919/aise.2005.000007},
   issn = {2097-5104},
   issue = {2},
   journal = {Artificial Intelligence Science and Engineering},
   month = {7},
   pages = {79-97},
   publisher = {Institute of Electrical and Electronics Engineers (IEEE)},
   title = {The Synergy of Seeing and Saying: Revolutionary Advances in Multi-modality Medical Vision-Language Large Models},
   volume = {1},
   year = {2025}
}

@article{Li2024,
   author = {Xiang Li and Congcong Wen and Yuan Hu and Zhenghang Yuan and Xiao Xiang Zhu},
   doi = {10.1109/MGRS.2024.3383473},
   issn = {21686831},
   issue = {2},
   journal = {IEEE Geoscience and Remote Sensing Magazine},
   pages = {32-66},
   publisher = {Institute of Electrical and Electronics Engineers Inc.},
   title = {Vision-Language Models in Remote Sensing: Current progress and future trends},
   volume = {12},
   year = {2024}
}

@misc{Xuyan2025,
   author = {Huang Xuyan and Sun Meng and Shen Chengxing and Li Haoxuan and Zhu Jianlin},
   doi = {10.1007/s00371-025-04109-y},
   issn = {01782789},
   issue = {13},
   journal = {Visual Computer},
   month = {10},
   pages = {11327-11348},
   publisher = {Springer Science and Business Media Deutschland GmbH},
   title = {Visual-language reasoning large language models for primary care: advancing clinical decision support through multimodal AI},
   volume = {41},
   year = {2025}
}

@inproceedings{Li2025,
  title={A Survey of State of the Art Large Vision Language Models: Benchmark Evaluations and Challenges},
  author={Li, Zongxia and Wu, Xiyang and Du, Hongyang and Liu, Fuxiao and Nghiem, Huy and Shi, Guangyao},
  booktitle={Proceedings of the Computer Vision and Pattern Recognition Conference},
  pages={1587--1606},
  year={2025}
}

@article{Ambrosini2010,
   author = {Dario Ambrosini and Claudia Daffara and Roberta Di Biase and Domenica Paoletti and Luca Pezzati and Roberto Bellucci and Francesca Bettini},
   doi = {10.1016/j.culher.2009.05.001},
   issn = {12962074},
   issue = {2},
   journal = {Journal of Cultural Heritage},
   pages = {196-204},
   publisher = {Elsevier Masson SAS},
   title = {Integrated reflectography and thermography for wooden paintings diagnostics},
   volume = {11},
   year = {2010}
}

@article{Zhang2017,
   author = {Hai Zhang and Stefano Sfarra and Karan Saluja and Jeroen Peeters and Julien Fleuret and Yuxia Duan and Henrique Fernandes and Nicolas Avdelidis and Clemente Ibarra-Castanedo and Xavier Maldague},
   doi = {10.1007/s10921-017-0414-8},
   issn = {15734862},
   issue = {2},
   journal = {Journal of Nondestructive Evaluation},
   month = {6},
   publisher = {Springer New York LLC},
   title = {Non-destructive Investigation of Paintings on Canvas by Continuous Wave Terahertz Imaging and Flash Thermography},
   volume = {36},
   year = {2017}
}

@article{Rippa2021,
   author = {Massimo Rippa and Vito Pagliarulo and Alessandra Lanzillo and Mariangela Grilli and Giancarlo Fatigati and Pasquale Rossi and Paola Cennamo and Giorgio Trojsi and Pietro Ferraro and Pasquale Mormile},
   doi = {10.1007/s10921-021-00755-z},
   issn = {15734862},
   issue = {1},
   journal = {Journal of Nondestructive Evaluation},
   month = {3},
   publisher = {Springer},
   title = {Active Thermography for Non-invasive Inspection of an Artwork on Poplar Panel: Novel Approach Using Principal Component Thermography and Absolute Thermal Contrast},
   volume = {40},
   year = {2021}
}

@article{Rippa2023,
   author = {M. Rippa and M. R. Vigorito and M. R. Russo and P. Mormile and G. Trojsi},
   doi = {10.1007/s10921-023-00972-8},
   issn = {15734862},
   issue = {3},
   journal = {Journal of Nondestructive Evaluation},
   month = {9},
   publisher = {Springer},
   title = {Active Thermography for Non-invasive Inspection of Wall Painting: Novel Approach Based on Thermal Recovery Maps},
   volume = {42},
   year = {2023}
}

@article{Peeters2018,
  title={IR reflectography and active thermography on artworks: the added value of the 1.5--3 $\mu$m band},
  author={Peeters, Jeroen and Van der Snickt, Geert and Sfarra, Stefano and Legrand, Stijn and Ibarra-Castanedo, Clemente and Janssens, Koen and Steenackers, Gunther},
  journal={Applied Sciences},
  volume={8},
  number={1},
  pages={50},
  year={2018},
  publisher={MDPI}
}

@misc{vlm,
      title={Qwen-VL: A Versatile Vision-Language Model for Understanding, Localization, Text Reading, and Beyond}, 
      author={Jinze Bai and Shuai Bai and Shusheng Yang and Shijie Wang and Sinan Tan and Peng Wang and Junyang Lin and Chang Zhou and Jingren Zhou},
      year={2023},
      eprint={2308.12966},
      archivePrefix={arXiv},
      primaryClass={cs.CV},
      url={https://arxiv.org/abs/2308.12966}, 
}

@article{Laureti2019,
   author = {Stefano Laureti and Hamed Malekmohammadi and Muhammad Khalid Rizwan and Pietro Burrascano and Stefano Sfarra and Miranda Mostacci and Marco Ricci},
   doi = {10.3390/s19194335},
   issn = {14248220},
   issue = {19},
   journal = {Sensors (Switzerland)},
   month = {10},
   pmid = {31597266},
   publisher = {MDPI AG},
   title = {Looking through paintings by combining hyper-spectral imaging and pulse-compression thermography},
   volume = {19},
   year = {2019}
}

@misc{Mercuri2011,
   author = {F. Mercuri and U. Zammit and N. Orazi and S. Paoloni and M. Marinelli and F. Scudieri},
   doi = {10.1007/s10973-011-1450-8},
   issn = {13886150},
   issue = {2},
   journal = {Journal of Thermal Analysis and Calorimetry},
   month = {5},
   pages = {475-485},
   title = {Active infrared thermography applied to the investigation of art and historic artefacts},
   volume = {104},
   year = {2011}
}

@article{ppt,
    author = {Maldague, X. and Marinetti, S.},
    title = {Pulse phase infrared thermography},
    journal = {Journal of Applied Physics},
    volume = {79},
    number = {5},
    pages = {2694-2698},
    year = {1996},
    month = {03},
    issn = {0021-8979},
    doi = {10.1063/1.362662},
    url = {https://doi.org/10.1063/1.362662},
}

@article{tsr1,
  title={Reconstruction and enhancement of active thermographic image sequences},
  author={Shepard, Steven M and Lhota, James R and Rubadeux, Bruce A and Wang, David and Ahmed, Tasdiq},
  journal={Optical Engineering},
  volume={42},
  number={5},
  pages={1337--1342},
  year={2003},
  publisher={Society of Photo-Optical Instrumentation Engineers}
}

@article{tsr2,
  title={Thermography of composites},
  author={Shepard, Steven M},
  journal={Materials Evaluation},
  volume={65},
  number={7},
  pages={690--696},
  year={2007}
}

@article{tsr3,
  title={The thermographic signal reconstruction method: A powerful tool for the enhancement of transient thermographic images},
  author={Balageas, Daniel L and Roche, Jean-Michel and Leroy, Fran{\c{c}}ois-Henri and Liu, Wei-Min and Gorbach, Alexander M},
  journal={Biocybernetics and biomedical engineering},
  volume={35},
  number={1},
  pages={1--9},
  year={2015},
  publisher={Elsevier}
}

@misc{salah_pt,
      title={Multi-Modal Attention Networks for Enhanced Segmentation and Depth Estimation of Subsurface Defects in Pulse Thermography}, 
      author={Mohammed Salah and Naoufel Werghi and Davor Svetinovic and Yusra Abdulrahman},
      year={2025},
      eprint={2501.09994},
      archivePrefix={arXiv},
      primaryClass={cs.CV},
      url={https://arxiv.org/abs/2501.09994}, 
}

@misc{salah_pca,
      title={PCA-Guided Autoencoding for Structured Dimensionality Reduction in Active Infrared Thermography}, 
      author={Mohammed Salah and Numan Saeed and Davor Svetinovic and Stefano Sfarra and Mohammed Omar and Yusra Abdulrahman},
      year={2025},
      eprint={2508.07773},
      archivePrefix={arXiv},
      primaryClass={eess.IV},
      url={https://arxiv.org/abs/2508.07773}, 
}

@article{Vavilov29112024,
author = {V.P Vavilov and P.G Bison and D.D Burleigh},
title = {Ermanno Grinzato’s contribution to infrared diagnostics and nondestructive testing: in memory of an outstanding researcher},
journal = {Quantitative InfraRed Thermography Journal},
volume = {21},
number = {6},
pages = {338--351},
year = {2024},
publisher = {Taylor \& Francis},
doi = {10.1080/17686733.2023.2170647},


URL = { 
    
        https://doi.org/10.1080/17686733.2023.2170647
    
    

},
eprint = { 
    
        https://doi.org/10.1080/17686733.2023.2170647
    
    

}

}

@article{Bison29112024,
author = {P. Bison and A. Bortolin and G. Cadelano and G. Ferrarini and M. Girotto and E. Guolo and F. Peron and M. Volinia},
title = {Ermanno Grinzato and the humidity assessment in porous building materials: retrospective and new achievements},
journal = {Quantitative InfraRed Thermography Journal},
volume = {21},
number = {6},
pages = {384--407},
year = {2024},
publisher = {Taylor \& Francis},
doi = {10.1080/17686733.2023.2231764},


URL = { 
    
        https://doi.org/10.1080/17686733.2023.2231764
    
    

},
eprint = { 
    
        https://doi.org/10.1080/17686733.2023.2231764
    
    

}

}

@article{RAJIC2002,
title = {Principal component thermography for flaw contrast enhancement and flaw depth characterisation in composite structures},
journal = {Composite Structures},
volume = {58},
number = {4},
pages = {521-528},
year = {2002},
issn = {0263-8223},
doi = {https://doi.org/10.1016/S0263-8223(02)00161-7},
author = {N Rajic}
}

@misc{Gavrilov2014,
   author = {D. Gavrilov and R. Gr Maev and D. P. Almond},
   doi = {10.1139/cjp-2013-0128},
   issn = {00084204},
   issue = {4},
   journal = {Canadian Journal of Physics},
   pages = {341-364},
   publisher = {National Research Council of Canada},
   title = {A review of imaging methods in analysis of works of art: Thermographic imaging method in art analysis},
   volume = {92},
   year = {2014}
}

@article{Chulkov2021,
   author = {A. O. Chulkov and S. Sfarra and N. Saeed and J. Peeters and C. Ibarra-Castanedo and G. Gargiulo and G. Steenackers and X. P.V. Maldague and M. A. Omar and V. Vavilov},
   doi = {10.1007/s10973-020-09326-2},
   issn = {15882926},
   issue = {5},
   journal = {Journal of Thermal Analysis and Calorimetry},
   month = {3},
   pages = {3835-3848},
   publisher = {Springer Science and Business Media B.V.},
   title = {Evaluating quality of marquetries by applying active IR thermography and advanced signal processing},
   volume = {143},
   year = {2021}
}
\end{document}